\documentclass{article}

\usepackage[preprint]{neurips_2025}

\usepackage[utf8]{inputenc} 
\usepackage[T1]{fontenc}    
\usepackage{hyperref}       
\usepackage{url}            
\usepackage{booktabs}       
\usepackage{amsfonts}       
\usepackage{nicefrac}       
\usepackage{microtype}      
\usepackage{xcolor}         
\usepackage{adjustbox}
\usepackage{blindtext}
\usepackage{multirow}
\usepackage{graphicx}
\usepackage{caption}
\usepackage{makecell}
\usepackage{subcaption}
\title{PhysDrive: A Multimodal Remote Physiological Measurement Dataset for In-vehicle Driver Monitoring}

\author{
Jiyao Wang$^{1*}$, Xiao Yang$^{1*}$, Qingyong Hu$^{2*}$, Jiankai Tang$^{3}$, Can Liu$^{4}$,  \\
\textbf{Dengbo He$^{1\dagger}$, Yuntao Wang$^{3}$, Yingcong Chen$^{1}$, Kaishun Wu$^{1}$} \\
$^{1}$The Hong Kong University of Science and Technology (Guangzhou) \\
$^{2}$The Hong Kong University of Science and Technology, $^{3}$Tsinghua University \\
$^{4}$Sichuan Agricultural University \\
\small{$*$ Equal Contribution \quad $\dagger$ Corresponding Author}
}

\begin{document}

\maketitle

\begin{abstract}
Robust and unobtrusive in-vehicle physiological monitoring is crucial for ensuring driving safety and user experience. While remote physiological measurement (RPM) offers a promising non-invasive solution, its translation to real-world driving scenarios is critically constrained by the scarcity of comprehensive datasets.  Existing resources are often limited in scale, modality diversity, the breadth of biometric annotations, and the range of captured conditions, thereby omitting inherent real-world challenges in driving. Here, we present PhysDrive, the first large-scale multimodal dataset for contactless in-vehicle physiological sensing with dedicated consideration on various modality settings and driving factors. PhysDrive collects data from 48 drivers, including synchronized RGB, near-infrared camera, and raw mmWave radar data, accompanied with six synchronized ground truths (ECG, BVP, Respiration, HR, RR, and SpO2). It covers a wide spectrum of naturalistic driving conditions, including driver motions, dynamic natural light, vehicle types, and road conditions. We extensively evaluate both signal‑processing and deep‑learning methods on PhysDrive, establishing a comprehensive benchmark across all modalities, and release full open‑source code with compatibility for mainstream public toolboxes. We envision PhysDrive will serve as a foundational resource and accelerate research on multimodal driver monitoring and smart‑cockpit systems.

\end{abstract}

\section{Introduction}

As intelligent transportation moves toward human–machine co-driving, effective driver monitoring becomes essential to ensure safety and enable timely driver intervention \citep{lu2023review,wang2024multi}. Such monitoring increasingly depends on understanding the driver's internal state, where physiological signals offer an objective and rich source of information for tasks like driver state monitoring \citep{wang2024efficient}, health assessment \citep{gharamohammadi2023vehicle}, and in-vehicle interaction \citep{detjen2021increase}. However, traditional contact-based acquisition methods for these signals, such as electrocardiography (ECG) and respiratory belts, are often intrusive and costly, potentially distracting drivers and hindering user acceptance \citep{yang2024video}. Remote physiological measurement (RPM) emerges as a compelling alternative, enabling the noninvasive, convenient, and simultaneous acquisition of multiple biological signals (e.g., heart rate (HR), respiration rate (RR)) without physical contact \citep{liang2024review}. Compared to contact-based approaches, RPM's noninvasive and convenient nature \citep{aarts2013non,yan2020high} facilitates its seamless integration into in-vehicle systems, thereby minimizing driver disruption and broadening its applicability within smart vehicles \citep{wang2024efficient}.

Contactless in-vehicle physiological monitoring primarily utilizes vision-based approaches and radio frequency (RF) sensing \citep{liang2024review}. Vision-based methods include cost-effective RGB cameras for color information-based remote photoplethysmography (rPPG) and near-infrared (NIR) cameras for more stable imaging than RGB under dynamic in-vehicle lighting. Millimeter-wave (mmWave) radar, as a typical RF technology, leverages its short wavelength to detect minute cardiorespiratory chest displacements at the millimeter level, offering robustness to illumination and enhanced privacy. While each modality has individual strengths, they also face distinct challenges: RGB is sensitive to light variations \citep{chiu2023deep}, NIR can have lower signal-to-noise ratios for some rPPG estimations \citep{verkruysse2008remote, vizbara2013comparison}, and mmWave systems can be influenced by vehicle vibrations and incur higher costs \citep{morabet2025driver, van2024deep}. 

Considering the variety of driving scenarios, it is essential to collaboratively analyze the three modalities towards a practical solution. Despite the growing interest in these technologies, the field is significantly constrained by a lack of comprehensive public benchmark datasets. As shown in Table \ref{dataset_summary}, existing datasets often focus on single-modality like RGB data \citep{tang2023mmpd, liu2024summit,stricker2014non, bobbia2019unsupervised} or provide limited coverage of NIR \citep{niu2019rhythmnet} or coarse-grained measurement for in-vehicle RF sensing \citep{li2024driver}. Besides, they are typically gathered in controlled indoor environments, which lack diversity in real-world settings.

In this paper, we propose PhysDrive, a multimodal non-intrusive dataset to facilitate the algorithm development for contactless driving physiological sensing. PhysDrive contains data from 48 drivers with over 1500k synchronized frames in total four conditions, from three contactless sensing modalities: RGB camera, NIR camera, and mmWave radar as well as six contact ground truths: ECG, blood volume pulse (BVP), respiration signals (RESP), HR, RR, and blood oxygen saturation (SpO2). To the best of our knowledge, PhysDrive is the first dataset that comprises all the modalities across real-world driving settings. The contributions and features of PhysDrive are as follows.

\textbf{Diverse Sensing Modalities.} PhysDrive aims to provide a comprehensive dataset considering the various applicability of existing sensors. It contains three typical contactless modalities in vision and RF sensors, with six contact ground truths as labeling across ECG, BVP, RESP, HR, RR, and SpO2. 

\textbf{Versatile Sensing under Real World Settings.} PhysDrive features practical data collection from 48 subjects under various real-world driving settings, such as different illuminations, motions, and road conditions. 
This opens up new research possibilities for establishing open evaluation benchmarks for in-vehicle RPM and unexplored problems, e.g., generalization in contactless multi-modal sensing.

\textbf{Extensive Benchmarks.} To demonstrate the utility of PhysDrive, we have extensively implemented and evaluated the performance of mainstream baseline models across all factors. We also provide open-source resources\footnote{https://github.com/WJULYW/PhysDrive-Dataset} to facilitate future research, including raw and preprocessed data, code for benchmarking setup, and tutorials for use with the public RPM toolbox.


\section{Related Works}
\subsection{Remote Physiological Measurement}
Effective driving monitoring systems are crucial for improving driver safety and well-being, with a growing demand for unobtrusive, contactless solutions \citep{wang2020unobtrusive, melders2025recent}. Among these, RGB cameras have attracted considerable attention, largely due to their ubiquity in vehicles. They mainly leverage the periodic color modulations in skin pixels induced by blood flow changes to perform remote photoplethysmography (rPPG) for estimating vital signs such as heart rate and respiratory rate \citep{poh2010advancements, speth2023non, wang2025physmle}. While promising, the efficacy of RGB-based rPPG is notably susceptible to fluctuations in ambient lighting conditions \citep{chiu2023deep}. To address this illumination challenge, near-infrared (NIR) cameras with active NIR illumination have been explored in the invisible light spectrum and share resilience to variations in visible light \citep{sun2024contrast, chen2018deepphys, yu2019remote}. However, due to the reduced sensitivity to blood oxygenation changes, it has low signal-to-noise ratio (SNR), which leads to higher requirements for algorithm development \citep{magdalena2018sparseppg}. Concurrently, RF-based solutions, such as mmWave radar, have been investigated from a different modality perspective. These systems typically transmit electromagnetic waves and analyze the reflected signals to capture mechanical movements associated with physiological processes, including chest vibrations from respiration and heartbeats. Thus, they inherently overcome challenges related to dynamic lighting conditions and can better preserve users' privacy compared to video collections. However, radar modules generally entail higher costs than camera systems \citep{liang2024review} and are susceptible to interference from vehicle vibrations during driving and diverse reflection paths from different vehicle interior structures \citep{morabet2025driver, van2024deep}. The distinct advantages and limitations of each sensing modality underscore the need for comprehensive, multi-modal datasets from realistic in-vehicle conditions. Our dataset fills this gap with synchronized modalities collected under diverse real-world scenarios to facilitate future research.. 

\begin{table}[!t]
\setlength{\tabcolsep}{1.5pt}
\caption{\small
Comparison of existing public remote physiological measurement datasets.}
\label{dataset_summary}

\begin{center}
\adjustbox{max width=\textwidth}{
\begin{tabular}{l c c l c l}
  \toprule[1.5pt]
  \textbf{Dataset} & \textbf{Subjects} & \textbf{Modalities} 
    & \textbf{Physiological Signals} & \textbf{Environment} & \textbf{Experiment Conditions} \\
  \midrule
  COHFACE \citep{heusch2017reproducible}       
    & 40  
    & \makecell[c]{RGB} 
    & BVP, RR                     
    & Indoor  
    & Illumination; Motion \\
  MAHNOB‐HCI \citep{soleymani2011multimodal}   
    & 27  
    & \makecell[c]{RGB} 
    & ECG, RR                     
    & Indoor  
    & Illumination; Emotion \\
  PURE \citep{stricker2014non}                 
    & 10  
    & \makecell[c]{RGB} 
    & BVP, HR, SpO2               
    & Indoor  
    & Motion \\
  UBFC‐rPPG \citep{bobbia2019unsupervised}      
    & 43  
    & \makecell[c]{RGB} 
    & BVP, HR, SpO2              
    & Indoor  
    & Motion \\
  VIPL‐HR \citep{niu2019rhythmnet}              
    & 107 
    & \makecell[c]{RGB, NIR} 
    & BVP, HR, SpO2               
    & Indoor  
    & Illumination; Skin color \\
  MMSE‐HR \citep{zhang2016multimodal}              
    & 140 
    & \makecell[c]{RGB} 
    & HR, RR, Blood pressure     
    & Indoor  
    & Skin color; Expression; Emotion \\
    HCW \citep{wang2025physmle}              
    & 48 
    & \makecell[c]{RGB} 
    & BVP, RESP, HR, RR     
    & Indoor  
    & Emotion \\
  ECG‐Fitness \citep{vspetlik2018visual}       
    & 17  
    & \makecell[c]{RGB} 
    & ECG, HR                     
    & Indoor  
    & Special state; Motion \\
  MMPD \citep{tang2023mmpd}                    
    & 33  
    & \makecell[c]{RGB} 
    & BVP, HR                     
    & Indoor  
    & Motion; Skin color; Illumination \\
      SUMS \citep{liu2024summit}                    
    & 10  
    & \makecell[c]{RGB} 
    & BVP, HR, RR, SpO2            
    & Indoor  
    & Motion \\
    iBVP \citep{joshi2024ibvp}                    
    & 32  
    & \makecell[c]{RGB, Thermal} 
    & BVP                     
    & Indoor  
    & Motion; Skin color \\
    SCAMP \citep{mcduff2022scamps}                    
    & 2800  
    & \makecell[c]{RGB} 
    & BVP, HR, RR, SpO2                 
    & Synthetics  
    & Illumination; Motion; Skin color\\
      SUMS \citep{liu2024summit}                    
    & 10  
    & \makecell[c]{RGB} 
    & BVP, HR, RR, SpO2            
    & Indoor  
    & Motion \\
  MMDE \citep{xiang2024multi}                  
    & 64  
    & \makecell[c]{RGB} 
    & BVP, HR                     
    & Indoor  
    & Illumination; Motion \\
  EquiPleth \citep{vilesov2022blending}         
    & 91  
    & \makecell[c]{RGB, mmWave} 
    & BVP                         
    & Indoor  
    & Body posture \\
  4TU.ResearchD \citep{sadeghi2024comprehensive}
    & 10  
    & \makecell[c]{mmWave} 
    & ECG                         
    & Indoor  
    & Angle; Special state \\
    MR‐NIRP \citep{nowara2020near}               
    & 18  
    & \makecell[c]{RGB, NIR} 
    & BVP                         
    & Driving  
    & Illumination \\
  Wu et al. \citep{wu2022compensation}          
    & 14  
    & \makecell[c]{RGB} 
    & HR                          
    & Driving  
    & Emotion; Illumination; Motion \\
  \midrule
  \textbf{PhysDrive}                                  
    & 48  
    & \makecell[c]{RGB, NIR,\\ mmWave} 
    & \makecell[l]{ECG, RESP, BVP,\\ HR, RR, SpO2} 
    & Driving 
    & \makecell[l]{Illumination; Motion; \\Road condition; Car type} \\
  \bottomrule[1.5pt]
\end{tabular}}
\end{center}
\vspace{-0.5cm}
\end{table}



\subsection{Multi-modal RPM Dataset}

\begin{figure*}
\begin{center}
\includegraphics[scale=0.28]{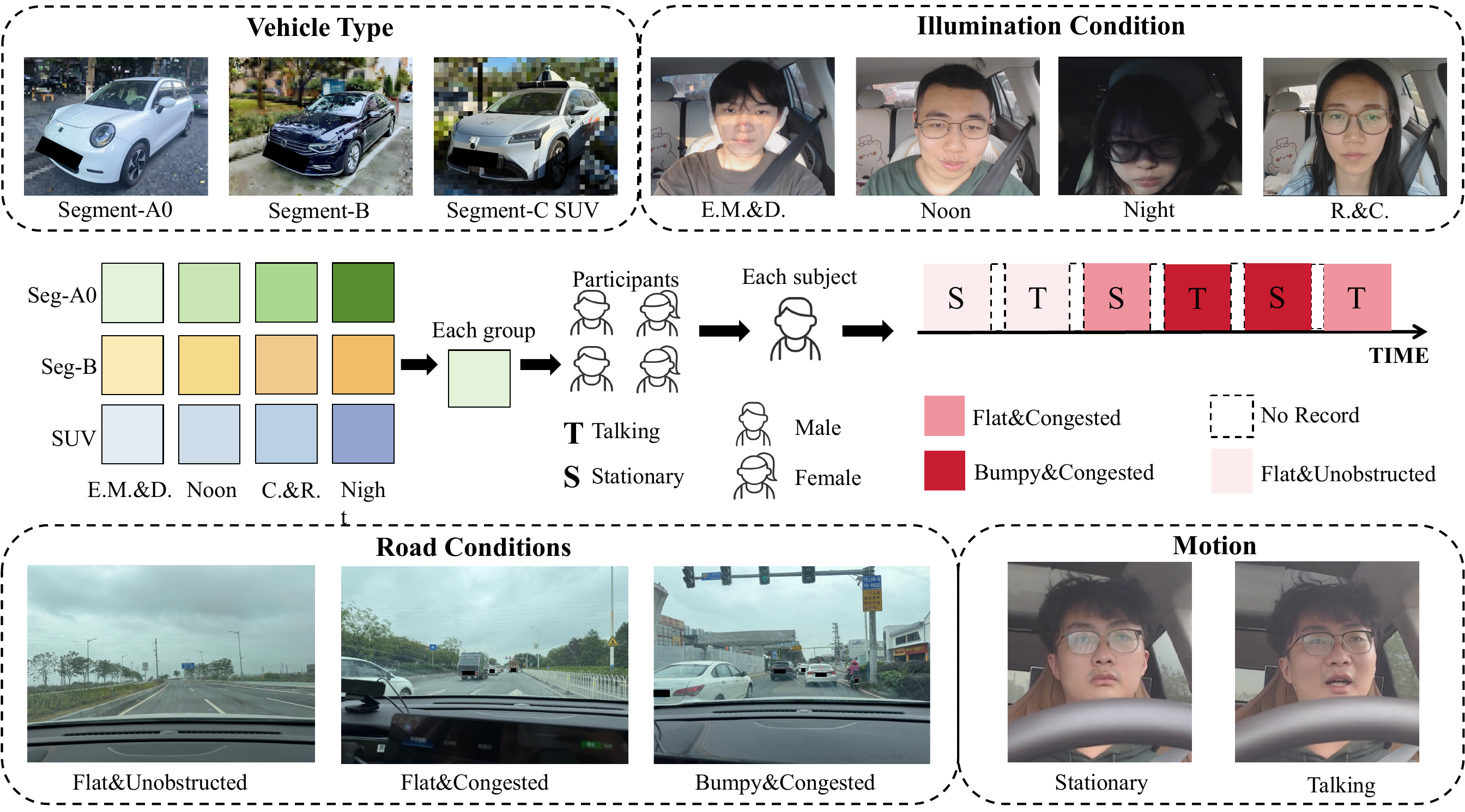}\\
\end{center}
\vspace{-2mm}
\caption{A visual illustration of our data collection experiment. Participants are divided into 12 groups by three types of vehicles and four types of illumination conditions. Each group consists of 4 subjects. Each subject data recordings of each participant are collected under three road conditions. In this figure, `E.M.\&D.' means `early morning and dusk', and `C.\&R.' is `cloudy and rainy day'. }\label{experiment_design}
\vspace{-5mm}
\end{figure*}

Currently, there are a number of datasets available for vision-based RPM; however, as indicated in Table \ref{experiment_design}, most of these datasets offer only a single modality (i.e., RGB video).The factors typically included in existing datasets focus on indoor motions, lighting conditions, and skin color. Some datasets also consider changes in human emotions \citep{soleymani2011multimodal, wang2025physmle, zhang2016multimodal} or various recording devices, such as webcams and mobile phone cameras \citep{niu2019rhythmnet,tang2023mmpd}. The algorithms developed and tested using these datasets are primarily suited for indoor monitoring scenarios instead of real driving scenarios. Only two datasets, MR-NIRP \citep{nowara2020near} and Wu et al. \citep{wu2022compensation}, concentrate on driving scenarios. Among these, MR-NIRP \citep{nowara2020near} is the only multimodal dataset specific to driving. Unfortunately, both datasets share common shortcomings: they are small in scale, provide only a single physiological label (either BVP or HR), and do not account for varying driving conditions or perceptual modalities.

Furthermore, there is a limited choice of raw mmWave data available. EquiPlet \citep{vilesov2022blending} and 4TU.ResearchD \citep{sadeghi2024comprehensive} are the only two public datasets that include mmWave data, with EquiPlet also offering RGB video. However, these datasets are focused on indoor environments and do not address in-vehicle monitoring. Since data collected indoors cannot effectively simulate the complex interactions of road surface feedback and signal reflections that occur during driving, such datasets are not suitable for developing and evaluating in-vehicle monitoring models with mmWave. Most previous mmWave-based in-vehicle monitoring systems \citep{juncen2023mmdrive, wang2021driver} have relied on private data, resulting in a lack of a public and unified evaluation benchmark.

For these reasons, we propose PhysDrive as the first public dataset that offers a comprehensive range of modal and physiological signal labels while encompassing the challenging factors found in real-world driving scenarios.

\section{Dataset}

\subsection{Data Collection} \label{Data Collection}

As mentioned earlier, dynamic lighting conditions and driver movements during driving can significantly impact the extraction of physiological signals from video data. Additionally, movements and road conditions pose challenges to the mmWave sensing method. To further investigate these factors, we designed a real-world driving experiment.

\textbf{Experiment Design.} Our data collection experiment was conducted in Guangzhou City, Guangdong Province, China. As shown in the Figure \ref{experiment_design}, we incorporated lighting conditions and vehicle types as between-subject factors in the experimental design. For the lighting conditions, we considered changes in angle and brightness under natural light. Specifically, we compared stable lighting conditions at \textit{Noon} (between 11 AM and 1 PM on a clear, cloudless day) to uneven light exposure and unstable angles experienced in the \textit{Early Morning} (7 AM to 9 AM) or \textit{Dusk} (5 PM to 7 PM) also on a clear day. Additionally, we accounted for two situations of varying light intensity: \textit{Rainy/Cloudy Days} and \textit{Nighttime} (after 8 PM). Regarding vehicle models, we considered the impact of different vehicle types on the mmWave scheme. We selected three common vehicle models according to the wheelbase classification method used by China: a \textit{A0-segment vehicle}, a \textit{B-segment vehicle}, and a \textit{C-segment SUV}. Each driver participated in only one type of vehicle under one specific lighting condition, ensuring that all drivers were evenly distributed among the 12 groups formed by the 3 vehicle models and 4 lighting conditions.

Furthermore, we treated the driver's actions and road conditions as within-subject variables. Each driver was required to navigate through a total of two action conditions and three road conditions. For the action variables, we defined two states: 1. \textit{Stationary State} - where the driver was instructed to avoid any distractions except for necessary head and hand movements for normal driving; and 2. \textit{Speaking State} - where we engaged the driver in conversation, encouraging them to perform additional safe actions, such as looking around and relaxing their shoulders. For the road condition variables, we defined three progressively difficult scenarios: 1. \textit{Flat and Unobstructed Road} - a newly constructed three-lane road with minimal traffic and no traffic lights; 2. \textit{Flat but Congested Road} - a three-lane road that, despite being flat, experiences heavy traffic, causing frequent stops and starts, leading to additional body movements; and 3. \textit{Bumpy and Congested Roads} - similar to the second condition but compounded by potholes, resulting in extra shaking. These factors together created six driving segments, each lasting approximately five minutes. 

All drivers were required to manually drive all six segments sequentially, as shown in Figure \ref{experiment_design}. It is important to note that the segments were not continuous, and we did not collect data for non-target driving segments between them, ensuring that each driver was recorded for six driving segments, about 30 minutes with 16.5 km driving distance in total.

\textbf{Participant.}  A total of 48 participants, aged from 18 to 41 (Mean=24.9, STD=4.1), were recruited through posters and word-of-mouth, with 24 females and 24 males, and were balanced within each group. All the participants are drivers who have obtained driving licenses for more than one year. Each driver will receive a compensation of 80 Chinese Yuan after completing the experiment. All participants were provided with information explaining the nature and purpose of the procedures involved in this study and signed the consent form before the start of the experiment. This research was approved by the Human and Artefacts Research Ethics Committee [HKUST(GZ)-HSP-2024-0076] of the Hong Kong University of Science and Technology (Guangzhou).


\subsection{Data Processing} \label{Data Processing}
\begin{figure*}
\begin{center}
\includegraphics[scale=0.21]{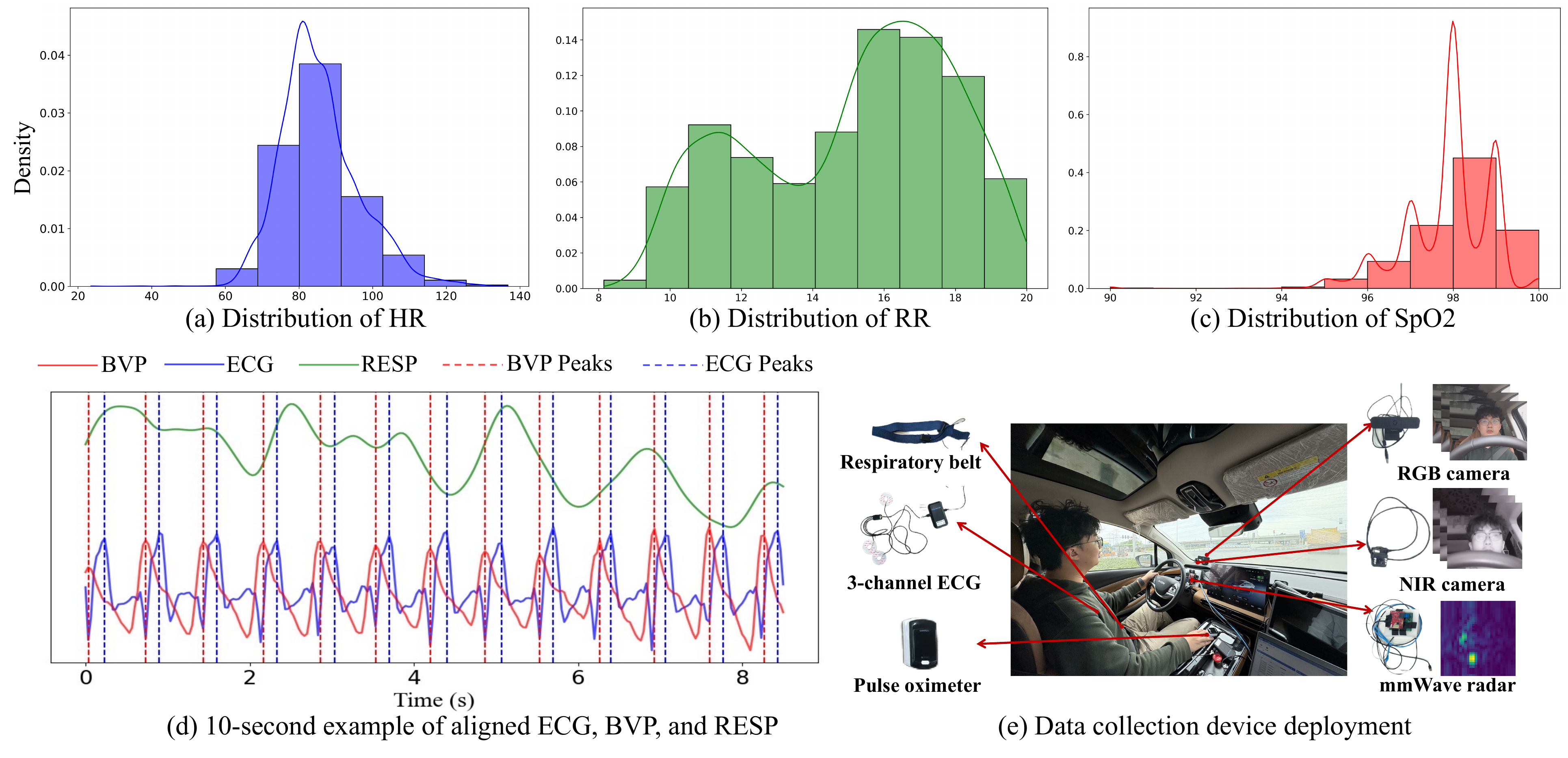}\\
\end{center}
\vspace{-3mm}
\caption{Visualization of processed physiological signals and deployment of data collection devices.}\label{distribution}
\vspace{-5mm}
\end{figure*}

\textbf{Apparatus.} For data acquisition, we utilize the PhysioLAB platform\footnote{https://www.infoinstruments.cn/product/physiolab/}, which is installed on our experimental laptop. This platform integrates a three-electrode ECG sensor, a respiratory belt transducer, a Logitech C925e RGB camera, and an NIR camera for synchronous data collection. The Ergoneers platform allows for the simultaneous recording and pausing of multiple devices, providing millisecond-level timestamps based on the laptop's local time. The acquisition frequency for physiological signals was set at 1000 Hz, while the RGB and IR video acquisition frequencies were set at 30 fps. RGB and IR video are recorded in `MP4' format, and ECG, RESP signals are in `CSV' file. For BVP and SpO2 data, we use the Contech CMS50E fingertip blood oxygen meter, which connects to the same laptop. The sampling frequencies are 60 Hz for BVP and 1 Hz for SpO2. The Contech CMS50E provides a second-level timestamp to synchronize the start of recording with the laptop.
The mmWave radar is configured as an effective bandwidth of 2.6 GHz with a virtual array of 12 antennas, with a range resolution of around 6 cm and an angular resolution of 14$^\circ$. It transmits 20 frames in one second. Each frame has 64 chirps with a velocity resolution of 7.6 cm/s.
The placement and installation positions of each device are shown in Figure \ref{distribution}(e).

\textbf{Data Synchronization.} To ensure robust temporal synchronization across the data acquisition platforms, all systems are hosted on a single laptop, and their respective timestamps are manually verified for consistency before starting each experimental session.
An ordered procedure for initiating and terminating recordings is implemented to further facilitate precise timestamp alignment. The PhysioLAB platform, chosen for its superior timestamp accuracy, initiates recording first and ends last, thereby encompassing the data streams from other sensors.
The Contech CMS50E, which provides timestamps accurate to the second, is started after PhysioLAB and stopped before it. Lastly, the mmWave radar system begins after the Contech device and is the first to stop recording.
The time interval between the start and end of recording across the different platforms typically does not exceed five seconds. Therefore, during the alignment of the collected data, we can ensure that the drift does not exceed one second. It can be seen from Figure \ref{distribution}(d) that, considering the naturally existing time shift between ECG and BVP, the signals we provide are well aligned.

Next, we take the three perceptual modalities as the benchmark and align them respectively with the physiological data. We align the RGB and NIR videos with ECG, RESP, BVP, and SpO2, and unify the aligned physiological signals to 30 Hz after up/down sampling. For mmWave data, we downsample the ECG and RESP signals to 20 Hz. 

\textbf{Data Preprocessing.} To verify the validity of the data and to prepare for the subsequent construction of the benchmark, we conducted preprocessing on the dataset. This dataset is intended for academic use only and is not allowed to be used for commercial purposes. Due to the privacy sensitivity of the original data and the storage limitations of the data hosting platform, we published the processed mmWave data of all participants, and the raw RGB and NIR data for one individual who agreed to publish without any data release agreement, along with the corresponding aligned physiological labels, on https://www.kaggle.com/datasets/xiaoyang274/physdrive. The way of accessing all of the raw data can be found at https://github.com/WJULYW/PhysDrive-Dataset.

For ECG, RESP, and BVP signals, the preprocessing steps mainly include bandpass filtering, signal credibility monitoring, and trend removal. Specifically, PhysDrive can be directly read and processed using rPPG-Toolbox~\citep{liu2023rppg}, enabling both intra-dataset and cross-dataset training. Inspired by iBVP~\citep{joshi2024ibvp}, we introduce a waveform similarity-based quality assessment method to enhance the preprocessing pipeline, allowing for more rigorous validation of signal integrity and completeness before model training. Further, HR and RR were derived using the NeuroKit2\footnote{https://neuropsychology.github.io/NeuroKit/} package from the ECG and RESP signals. The distribution of three indicators and one visualization example of processed physiological signals is shown in Figure \ref{distribution}. It is worth noting that due to safety considerations in our experiment, induction for low SpO2 was not carried out, so the distribution of SpO2 was relatively concentrated. Therefore, our subsequent benchmarks will not be targeted at SpO2.

For RGB and NIR video, the resolution of the raw RGB and NIR frames is 1024×576 and 1920x1080, respectively. In order to meet the input requirements of most baselines and reduce the reading time during the model training process, we follow \citep{niu2019rhythmnet, yu2023physformer++}, first locate the face, and then crop all frames to a size of 128x128 and enlarge the face area. We take it as the input of the baseline for directly inputting the video. Further, we generate the frames of the cropped images as STMap as the input of some baselines according to Appendix \ref{app:a}. 

The raw mmWave data is first organized as a tensor $X \in \mathbb{R}^{N_f\times2 \times N_r \times N_a \times N_d}$ with range-angle-doppler FFT operations. Here, $N_f$ represents the number of time frames, the dimension of size `2' accounts for the real and imaginary components of the complex radar signals, and $N_r, N_a, N_d$ denote the number of bins partitioning the signal by range (distance to reflectors), angle (spatial direction of reflections), and Doppler (velocity of reflectors), respectively. To isolate subtle dynamic physiological signals, we first apply static reflection removal techniques to filter out interference from stationary objects within the environment and then localize the human subject \citep{hu2024contactless}. Subsequently, to enhance computational efficiency and concentrate on the subject's immediate vicinity, we crop the data around the localized region. This results in a focused tensor with $N_r=8$ range bins (a physical range of 48 cm to cover typical human torso), $N_a=16$ angle bins (45° angle coverage, to cover frontal human presence), and $N_d=8$ Doppler bins (a velocity range of 60.8 cm/s on cardiorespiratory motions without gross body movements). Finally, we segment the data into sequences of $N_f=200$ frames, equivalent to 10 seconds of observation, ensuring temporal alignment with other sensor modalities.



\section{Benchmarks} \label{Benchmarks}
\subsection{Benchmark Setup} \label{Benchmark Setup}

\textbf{Baselines.} We choose five non-learning RGB-based traditional methods: CHROM \citep{de2013robust}, POS \citep{wang2016algorithmic}, GREEN \citep{verkruysse2008remote}, ICA \citep{poh2010advancements}, and ARM-RR \citep{tarassenko2014non}. Besides, we select two unsupervised DL methods: Contrast-Phys+ \citep{sun2024contrast} and SiNC \citep{speth2023non}, which are trained on unlabelled PhysDrive RGB video and tested on the test set. Other seven single-task DL methods (including DeepPhys \citep{chen2018deepphys}, PhysNet \citep{yu2019remote}, PhysFormer \citep{yu2022physformer}, EfficientPhys \citep{liu2023efficientphys}, Rhythmformer \citep{zou2024rhythmformer}, BVPNet \citep{das2021bvpnet}, RhythmNet \citep{niu2019rhythmnet}) and four multi-task learning (MTL) baselines (including MTTS-CAN \citep{liu2020multi}, BigSmall \citep{narayanswamy2024bigsmall}, BaseNet \citep{wang2025physmle}, PhysMLE \citep{wang2025physmle}). Among them, Contrast-Phys+ \citep{sun2024contrast}, DeepPhys \citep{chen2018deepphys}, and PhysNet \citep{yu2019remote} also serve as the baselines for NIR sensing. For mmWave sensing, we evaluate three baselines (including IQ-MVED \citep{zheng2021more}, VitaNet \citep{khan2022contactless}, and mmFormer \citep{hu2024contactless}). Note that, since we didn't find the proper MTL mmWave method, we make minor changes to the above method, including: (1) aggregating multiple Doppler bins with a linear layer; (2) adding new estimation heads after the backbone network. Besides, we remove the AU head from BigSmall for fair comparison. RhythmNet, BaseNet, and PhysMLE take STMap as input.

\textbf{Implementation.} All models used in the evaluation are implemented in PyTorch. The implementation of the baselines is primarily sourced from the rPPG-toolbox \citep{liu2023rppg}, otherwise from their open-source repositories. Our experiments are conducted on two servers, one equipped with 8 Nvidia RTX 3090 cards and the other with 8 Nvidia RTX A6000 cards. For the parameter settings of each baseline, we adhere to the descriptions provided in their respective source papers or the default configurations found in the rPPG-toolbox. 

\textbf{Metrics. }The goal of this dataset is to study RPM. Depending on our different sensing methods, we have outlined the following specific tasks: For the RGB and NIR sensing methods, we aim to fit the BVP and RESP signals, while evaluating HR and RR extracted from BVP and RESP, respectively. 
The mmWave methods in our task focus on HR and RR values regression, as well as ECG and RESP signals recovery, simultaneously.
We have adopted evaluation metrics from previous studies \citep{wang2024generalizable,wang2024condiff}, including mean absolute error (MAE), root mean square error (RMSE), and Pearson’s correlation coefficient (P) to assess the estimated indicators.

\subsection{Cross-subject Evaluation}

\textbf{Evaluation Protocol.}  To evaluate the dataset's usability, we first conduct an intra-dataset cross-subject evaluation within PhysDrive. We utilize data from 48 drivers for training, validation, and testing in a distribution of 80\%, 10\%, and 10\%, respectively. By controlling random seeds, we perform five independent evaluations, and we report the average values of all indicators across these five evaluations. 

\begin{table*}[!t]
\addtocounter{table}{-1}  
  \setlength{\tabcolsep}{3pt}
  \small
  \centering
  \begin{subtable}[t]{0.48\textwidth}
    \refstepcounter{table}\label{cross_subject_single}
    \centering
    \caption{\small 
      Table 2: \textbf{Intra-dataset.} HR estimation performance on RGB, NIR, and mmWave modalities.}
    \adjustbox{max width=\textwidth}{
      \begin{tabular}{l l c c c}
        \toprule[1.5pt]
        \textbf{Modality} & \textbf{Method} & MAE$\downarrow$ & RMSE$\downarrow$ & P$\uparrow$ \\
        \midrule
        \parbox[t]{2mm}{\multirow{6}{*}{RGB}}
        & CHROM \citep{de2013robust}         & 12.23 & 15.97 & 0.11 \\
        & POS \citep{wang2016algorithmic}     & 12.42 & 16.15 & 0.10 \\
        & SiNC \citep{speth2023non}           & 13.49 & 16.57 & 0.03 \\
        & PhysNet \citep{yu2019remote}        &  6.29 &  8.93 & 0.61 \\
        & RhythmFormer \citep{zou2024rhythmformer} &  7.21 &  9.84 & 0.45 \\
        & RhythmNet$^{*}$ \citep{niu2019rhythmnet} &  6.84 &  9.01 & 0.58 \\
        \midrule
        \parbox[t]{2mm}{\multirow{2}{*}{NIR}}
        & PhysNet \citep{yu2019remote}        & 10.69 & 13.21 & 0.12 \\
        & Contrast-Phys+ \citep{sun2024contrast} & 13.65 & 16.08 & 0.05 \\
        \midrule
        \parbox[t]{2mm}{\multirow{2}{*}{mmWave}}
        & VitaNet \citep{khan2022contactless}    &  4.94 &  7.15 & 0.94 \\
        & mmFormer \citep{hu2024contactless}   & 3.65  & 5.09  & 0.97 \\
        \bottomrule[1.5pt]
      \end{tabular}}
    \\[1pt]
  \end{subtable}\hfill
  \begin{subtable}[t]{0.48\textwidth}
    \refstepcounter{table}\label{cross_subject_mtl}
    \centering
    \caption{\small 
      Table 3: \textbf{Intra-dataset.} Multi-task estimation performance of MTL methods.}
    \adjustbox{max width=\textwidth}{
      \begin{tabular}{l c c c c c c}
        \toprule[1.5pt]
        \multirow{2}{*}{\textbf{Method}} & \multicolumn{3}{c}{\textbf{HR}} & \multicolumn{3}{c}{\textbf{RR}} \\
        \cmidrule(lr){2-4} \cmidrule(lr){5-7}
        & MAE$\downarrow$ & RMSE$\downarrow$ & P$\uparrow$ 
        & MAE$\downarrow$ & RMSE$\downarrow$ & P$\uparrow$ \\
        \midrule
        CHROM \citep{de2013robust}        & 12.23 & 15.97 & 0.11 & N/A   & N/A   & N/A \\
        ARM-RR \citep{tarassenko2014non}  & N/A   & N/A   & N/A  & 4.63  & 5.88  & 0.08 \\
        MTTS-CAN \citep{liu2020multi}     & 8.75  & 11.02 & 0.26 & 3.01  & 4.14  & 0.12 \\
        BigSmall \citep{narayanswamy2024bigsmall} & 9.21  & 11.57 & 0.24 & 3.18  & 4.29  & 0.10 \\
        BaseNet$^{*}$ \citep{wang2025physmle}    & 6.97  & 9.32  & 0.53 & 2.70  & 3.29  & 0.15 \\
        PhysMLE$^{*}$ \citep{wang2025physmle}    & 7.02  & 9.90  & 0.55 & 2.21  & 3.59  & 0.16 \\
        IQ-MVED \citep{zheng2021more}     & 14.17 & 18.21 & 0.45 & 3.71  & 4.69  & 0.04 \\
        \ \ -(Wave Recovery)  & 29.18 & 37.31 & 0.10 & 2.20  & 3.15  & 0.32 \\
        VitaNet \citep{khan2022contactless}     & 4.94  & 7.15  & 0.94 & 2.56  & 3.64  & 0.80 \\
        \ \ -(Wave Recovery)  & 35.21 & 43.1 & 0.07 & 2.88  & 3.71  & 0.12 \\
        mmFormer \citep{hu2024contactless}    & 3.65  & 5.09  & 0.97 & 1.49  & 2.41  & 0.83 \\
        \ \ -(Wave Recovery)  & 33.87 & 41.96 & 0.03 & 2.1  & 3.12  & 0.37 \\
        \bottomrule[1.5pt]
      \end{tabular}}\\[1pt]
  \end{subtable}
  \vspace{-0.5cm}
\end{table*}

\textbf{Results of HR Estimation.}  We present main findings in Table \ref{cross_subject_single}, and the complete version can be found in Table \ref{app:intra_full} in Appendix \ref{app:b}. As shown in Table \ref{cross_subject_single}, from the perspective of data validity, our results align closely with those from previous similar datasets. The baseline methods using RGB, NIR, and mmWave modalities achieve reasonable results on our dataset, particularly the DL method. This indicates that our dataset effectively validates the input modalities, such as physiological signals (\textit{i.e.}, ECG and BVP), and the correlations between inputs and outputs.

Regarding the baseline results across multiple modalities, it's noteworthy that RGB techniques significantly outperform the NIR methods. This finding is consistent with results reported in \citep{nowara2020near}. However, the vision-based solutions do not match the accuracy levels achieved in the indoor rPPG dataset under simpler protocols, such as intra-dataset evaluations. For example, results from \citep{sun2024contrast} indicate that PhysNet achieves a MAE of 2.1 and a P of 0.99 in intra-dataset evaluations on PURE. Furthermore, the mmWave methods substantially outperform all vision-based approaches, particularly in terms of P value. For instance, VitaNet exceeded the best RGB method (PhysNet) by about 54.1\%.

\textbf{Results of Multi-task Estimation.}  In Table \ref{cross_subject_mtl}, we evaluate the performance of the MTL methods of three modalities. The results demonstrate not only the validity of HR-related BVP and ECG signals but also confirm the correlation between these inputs and RESP signals in the context of RR monitoring. In terms of baseline performance, we observe that among vision-based methods, MTL techniques built on STMap (i.e., BaseNet and PhysMLE) outperform those that directly use face videos. Notably, for the RR estimation task, the average P value improved by approximately 25\%. 
While mmWave methods show strengths, their performance notably drops for HR/RR extraction via recovered ECG/RESP waveforms compared to direct parameter regression. This underperformance is plausibly attributed to mmWave's acute sensitivity to temporal misalignment during the critical waveform regression step. Consequently, a key research direction is to develop training schemes that confer temporal robustness to mmWave models, thereby alleviating stringent data synchronization demands and potentially broadening practical applicability.

\textbf{Results of Cross-scenario Evaluation.}  To address the two challenging factors that affect the RPM methods (\textit{i.e.,} driver motions and road conditions), we perform cross-scenario evaluations, assessing data from different segments of the test set individually. The results are displayed in Table \ref{intra_scenario}. 

Firstly, as shown in Table \ref{intra_scenario}, aligned with Table \ref{cross_subject_mtl}, mmWave methods outperform RGB and NIR methods, while the impact of motions is larger on mmWave methods. Specifically, regarding driver actions, as expected, the performance of all methods in both HR and RR monitoring tasks declines when drivers are in the talking state compared to the stationary state.

Similarly, seeing Table \ref{intra_scenario}, as road conditions more challenging due to jolting and congestion, the accuracy of the mmWave solution decreases. Specifically, compared to flat and unobstructed roads, the performance on more congested flat roads experience a minor decline, usually not exceeding 5\%. However, when the road surface becomes bumpy, the degradation is more pronounced, with the MAE increasing by about 13\% compared to flat and unobstructed roads.

\begin{table*}[!t]
\setlength{\tabcolsep}{1.5pt}
\caption{\small
\textbf{Intra‐dataset Scenario Evaluation.} HR and RR estimation performances under driver motion conditions and road scenarios. PhysMLE: RGB, PhysNet: NIR, VitaNet and mmFormer: mmWave.}
\label{intra_scenario}
\scriptsize
\begin{center}
\adjustbox{max width=\textwidth}{
\begin{tabular}{l *{5}{c c c c}}
\toprule[1.5pt]
\multirow{3}{*}{\textbf{Method}}
  & \multicolumn{4}{c}{\textbf{Stationary}}
  & \multicolumn{4}{c}{\textbf{Talking}}
  & \multicolumn{4}{c}{\textbf{Flat \& Unobstructed}}
  & \multicolumn{4}{c}{\textbf{Flat \& Congested}}
  & \multicolumn{4}{c}{\textbf{Bumpy \& Congested}} \\
\cmidrule(lr){2-5}\cmidrule(lr){6-9}\cmidrule(lr){10-13}\cmidrule(lr){14-17}\cmidrule(lr){18-21}
  & \multicolumn{2}{c}{\textbf{HR}} & \multicolumn{2}{c}{\textbf{RR}}
  & \multicolumn{2}{c}{\textbf{HR}} & \multicolumn{2}{c}{\textbf{RR}}
  & \multicolumn{2}{c}{\textbf{HR}} & \multicolumn{2}{c}{\textbf{RR}}
  & \multicolumn{2}{c}{\textbf{HR}} & \multicolumn{2}{c}{\textbf{RR}}
  & \multicolumn{2}{c}{\textbf{HR}} & \multicolumn{2}{c}{\textbf{RR}} \\
\cmidrule(lr){2-3}\cmidrule(lr){4-5}
\cmidrule(lr){6-7}\cmidrule(lr){8-9}
\cmidrule(lr){10-11}\cmidrule(lr){12-13}
\cmidrule(lr){14-15}\cmidrule(lr){16-17}
\cmidrule(lr){18-19}\cmidrule(lr){20-21}
  & MAE$\downarrow$ & P$\uparrow$ & MAE$\downarrow$ & P$\uparrow$
  & MAE$\downarrow$ & P$\uparrow$ & MAE$\downarrow$ & P$\uparrow$
  & MAE$\downarrow$ & P$\uparrow$ & MAE$\downarrow$ & P$\uparrow$
  & MAE$\downarrow$ & P$\uparrow$ & MAE$\downarrow$ & P$\uparrow$
  & MAE$\downarrow$ & P$\uparrow$ & MAE$\downarrow$ & P$\uparrow$ \\
\midrule
PhysMLE \citep{wang2025physmle}
  & 6.55 & 0.58 & 2.12 & 0.18
  & 6.98 & 0.54 & 2.30 & 0.15
  & 6.95 & 0.56 & 2.16 & 0.17
  & 7.01 & 0.55 & 2.22 & 0.15
  & 7.16 & 0.53 & 2.32 & 0.14 \\
PhysNet (NIR) \citep{yu2019remote}
  &10.35 & 0.15 &  N/A &  N/A
  &11.13 & 0.11 &  N/A &  N/A
  &10.59 & 0.14 &  N/A &  N/A
  &10.67 & 0.12 &  N/A &  N/A
  &10.73 & 0.11 &  N/A &  N/A \\
VitaNet \citep{khan2022contactless}
  & 4.92 & 0.93 & 2.47 & 0.80
  & 4.99 & 0.92 & 2.80 & 0.77
  & 4.87 & 0.94 & 2.44 & 0.81
  & 5.17 & 0.92 & 2.56 & 0.80
  & 5.97 & 0.90 & 2.86 & 0.78 \\
mmFormer \citep{hu2024contactless}
  & 3.32 & 0.97 & 1.24 & 0.86
  & 3.75 & 0.94 & 1.51 & 0.84
  & 3.32 & 0.97 & 1.43 & 0.85
  & 3.39 & 0.97 & 1.51 & 0.86
  & 3.77 & 0.94 & 1.55 & 0.84 \\
\bottomrule[1.5pt]
\end{tabular}}
\end{center}
\vspace{-0.3cm}
\end{table*}

\subsection{Cross-dataset Evaluation}
\textbf{Evaluation Protocol.}  To assess the capability of our dataset as both a training and a test set for evaluating the generalization ability of the DL model, we conduct extensive evaluations based on a cross-dataset protocol. We select the PURE \citep{stricker2014non} and UBFC-rPPG \citep{bobbia2019unsupervised} datasets, which are commonly used in previous studies \citep{wang2024generalizable,yu2023physformer++}, alongside PhysDrive for our evaluation. It is important to note that neither PURE nor UBFC-rPPG provides RESP, and neither PURE nor PhysDrive is specifically designed for measuring blood oxygen levels. Thus, our evaluations focus solely on HR estimation tasks. The main results are shown in Table \ref{cross_scenario_main}, while the remaining results can be referenced in Table \ref{cross_scenario_traditional}, \ref{cross_scenario} in Appendix \ref{Cross-dataset Evaluation: Tested on Other Datasets}, \ref{Cross-dataset Evaluation on Different Scenarios}.

\begin{table*}[!t]
\setlength{\tabcolsep}{1.5pt}
\caption{\small
\textbf{Cross‐dataset Scenario Evaluation.} HR estimation performance of RGB‐based methods under varying lighting and motion conditions when trained on PURE and UBFC‐rPPG. Here, `E.M.\&D.' indicates early morning\&dusk; and `R.\&C.' is rainy\&cloudy.}
\label{cross_scenario_main}
\scriptsize
\begin{center}
\adjustbox{max width=\textwidth}{
\begin{tabular}{l l *{7}{c c}}
  \toprule[1.5pt]
  \multirow{2}{*}{\textbf{Method}} & \multirow{2}{*}{\textbf{Train Set}}
    & \multicolumn{2}{c}{\textbf{E.M.\&D.}} 
    & \multicolumn{2}{c}{\textbf{Noon}} 
    & \multicolumn{2}{c}{\textbf{Night}} 
    & \multicolumn{2}{c}{\textbf{R.\&C.}} 
    & \multicolumn{2}{c}{\textbf{Stationary}} 
    & \multicolumn{2}{c}{\textbf{Talking}} 
    & \multicolumn{2}{c}{\textbf{All}} \\
  \cmidrule(lr){3-4}\cmidrule(lr){5-6}\cmidrule(lr){7-8}
  \cmidrule(lr){9-10}\cmidrule(lr){11-12}\cmidrule(lr){13-14}\cmidrule(lr){15-16}
  & 
    & MAE$\downarrow$ & P$\uparrow$ 
    & MAE$\downarrow$ & P$\uparrow$ 
    & MAE$\downarrow$ & P$\uparrow$ 
    & MAE$\downarrow$ & P$\uparrow$ 
    & MAE$\downarrow$ & P$\uparrow$ 
    & MAE$\downarrow$ & P$\uparrow$ 
    & MAE$\downarrow$ & P$\uparrow$ \\
  \midrule
  \multirow{2}{*}{SiNC \citep{speth2023non}}
    & PURE       & 11.16 & 0.06 & 10.06 & 0.17 & 13.24 & 0.01 & 11.39 & 0.11 & 11.90 & 0.11 & 11.02 & 0.13 & 11.46 & 0.11 \\
    & UBFC‐rPPG  & 15.99 & 0.03 & 11.62 & 0.26 & 15.42 & -0.01 & 13.99 & 0.08 & 14.07 & 0.14 & 14.39 & 0.08 & 14.26 & 0.12 \\
  \midrule
  \multirow{2}{*}{RhythmFormer \citep{zou2024rhythmformer}}
    & PURE       & 12.51 & 0.04 & 10.40 & 0.19 & 13.65 & -0.05 & 11.73 & 0.04 & 11.45 & 0.13 & 12.09 & 0.07 & 11.72 & 0.11 \\
    & UBFC‐rPPG  & 12.98 & 0.01 & 11.04 & 0.15 & 12.94 & 0.05 & 12.18 & -0.01 & 12.16 & 0.11 & 12.57 & 0.05 & 12.53 & 0.08 \\
  \midrule
  \multirow{2}{*}{RhythmNet$^*$ \citep{niu2019rhythmnet}}
    & PURE       &  7.91 & 0.04 &  8.87 & 0.23 & 11.96 & -0.06 &  8.40 & 0.14 &  9.61 & 0.15 &  8.99 & 0.11 &  9.30 & 0.14 \\
    & UBFC‐rPPG  &  6.83 & 0.03 & 10.56 & 0.15 & 12.25 & -0.03 &  9.95 & 0.01 & 10.18 & 0.11 &  9.56 & 0.17 &  9.86 & 0.14 \\
  \bottomrule[1.5pt]
\end{tabular}} \\
\end{center}
\vspace{-0.5cm}
\end{table*}


\textbf{Results of Cross-scenario Evaluation.}  We evaluate methods trained on other datasets across various challenging scenarios in PhysDrive. As shown in Table \ref{cross_scenario_main}, we can conclude that, first, all methods achieve the best performance around noon and the lowest performance at night. Second, conditions on cloudy and rainy days, which have varying brightness but maintain stability, outperform those in the early morning and at dusk, when lighting fluctuates with the driving direction. As expected, MAE may vary when drivers engage in additional conversational behaviors.

\section{Discussion and Limitations}\label{Discussion and Limitations}

In this paper, we present a novel multimodal in-vehicle driver RPM dataset called PhysDrive. Extensive evaluations have confirmed the validity of the data, revealing some interesting findings.

\textbf{Modality Comparison.}  Our evaluation of RGB, NIR, and mmWave modalities reveals distinct strengths that depend on the modality and scenario. In almost all driving conditions, mmWave radar methods demonstrate superior accuracy in directly estimating indicators (i.e., HR, RR) and recovering RESP. However, they consistently struggle to reconstruct ECG waveforms. This discrepancy likely arises because ECG captures detailed cardiac features \citep{lu2009comparison}, that are inherently more difficult to recover and require stricter time synchronization than BVP signals. Vision-based methods excel at recovering BVP waveforms, particularly for RGB methods with the highest SNR in well-lit conditions. Although NIR methods are theoretically more stable than RGB in low-light conditions, their performances do not exceed that of RGB due to a lower SNR as aligned with \citep{nowara2020near}. 

Our analyses show that the smoothness of the road surface has the greatest effect on mmWave's performance, followed by driver movements and traffic congestion. In contrast, vision-based methods are particularly sensitive to brightness and quick changes in lighting. This suggests the possibility of dynamic modality selection or adaptive fusion to leverage each sensor's strengths effectively \citep{kurihara2021non,vilesov2022blending}.

\textbf{Insights for RPM Models.}  Compared with traditional signal-processing baselines, fully supervised deep-learning models, and unsupervised (or self-supervised) approaches under the same evaluation protocols, distinct training paradigms emerge. 
Supervised DL models, while effective in-domain, often exhibit poor generalization across different vehicle, lighting, or road scenarios, likely due to label noise and environmental overfitting. 
Conversely, unsupervised pretraining on large volumes of unlabeled driving data leads to representations that generalize more robustly across datasets. Therefore, we recommend a two-stage training strategy: initially, learn robust feature extractors using unsupervised/self-supervised methods on diverse, unlabeled real-world driving videos, followed by supervised fine-tuning on annotated data tailored to the specific application context.

Architecturally, we have observed that STMap methods for RGB video outperform direct video-input networks in multi-task estimation and generalization. By converting subtle changes in facial color into structured maps, STMap reduces the effects of noise from head motion and variations in illumination. However, this preprocessing pipeline introduces additional latency, which may hinder real-time applications. Future research should explore learnable or more efficient preprocessing techniques that maintain the robustness of STMap while minimizing computational overhead.

\textbf{Future Works Enabled by PhysDrive. }The PhysDrive dataset supports future research in remote physiological monitoring and intelligent in-vehicle systems. Its rich multimodal data encourages the development of sensor fusion techniques and advanced representation learning for robust physiological measurement across diverse driving conditions. For mmWave sensing, PhysDrive's provision of raw data will be invaluable for creating training schemes that enhance temporal robustness, potentially reducing the strictness of synchronization requirements in practical data collection.

\textbf{Limitations.} Despite its comprehensiveness, the PhysDrive dataset still has several limitations. First, our participant cohort predominantly consists of individuals of East Asian descent, which limits the evaluation of camera-based methods across a diverse range of skin tones \citep{tang2023mmpd}. Future data collection should prioritize the inclusion of individuals with different skin tones to assess and mitigate rPPG biases. Secondly, we currently focus on drivers as the first step. We envision the principles within PhysDrive can inspire initial explorations into passenger monitoring, adapting models to new challenges such as varied seating positions and occlusions \citep{van2024deep}.
Third, although we have SpO2 recordings, the lack of a dedicated acquisition protocol results in limited variability, diminishing its utility. Targeted experiments, such as those under simulated hypoxia or high-altitude conditions, are necessary to develop reliable in-vehicle SpO2 estimation methods. Finally, the current one-second synchronization tolerance between modalities may obscure transient cardiac events; future datasets should implement hardware-level timestamping or higher-precision synchronization to support analyses of rapid physiological changes.

\section{Conclusion}
In this paper, we introduce PhysDrive, a multimodal dataset designed for the RPM of in-vehicle drivers.
PhysDrive is the first to focus on and meticulously design driving scenarios, simultaneously providing RGB, NIR, and mmWave sensing data as well as a large-scale dataset of various physiological signals (including ECG, RESP, BVP, HR, RR, and SpO2). It aims to meet the demand for a relatively complete public dataset in the previous research communities of different monitoring technologies. The designs of various scenarios provided in PhysDrive can offer a relatively comprehensive public benchmark for subsequent work. It also provides an opportunity to integrate multiple sensing technologies and communities, accelerating the development of the next-generation intelligent cockpit.




\bibliographystyle{plainnat}
\bibliography{neurips_2025}

\clearpage


\appendix

\section{Producing Spatial-Temporal Map from RGB Video} \label{app:a}
To reduce the computational resources required during the training process and encourage the model to focus more on facial skin brightness changes rather than environmental variations, we adopted the ROI extraction method from \citep{niu2019rhythmnet} and transformed the video into a Spatial-Temporal Map (STMap) to provide a more lightweight input format. 

\begin{figure*}[h]
\begin{center}
\includegraphics[scale=0.4]{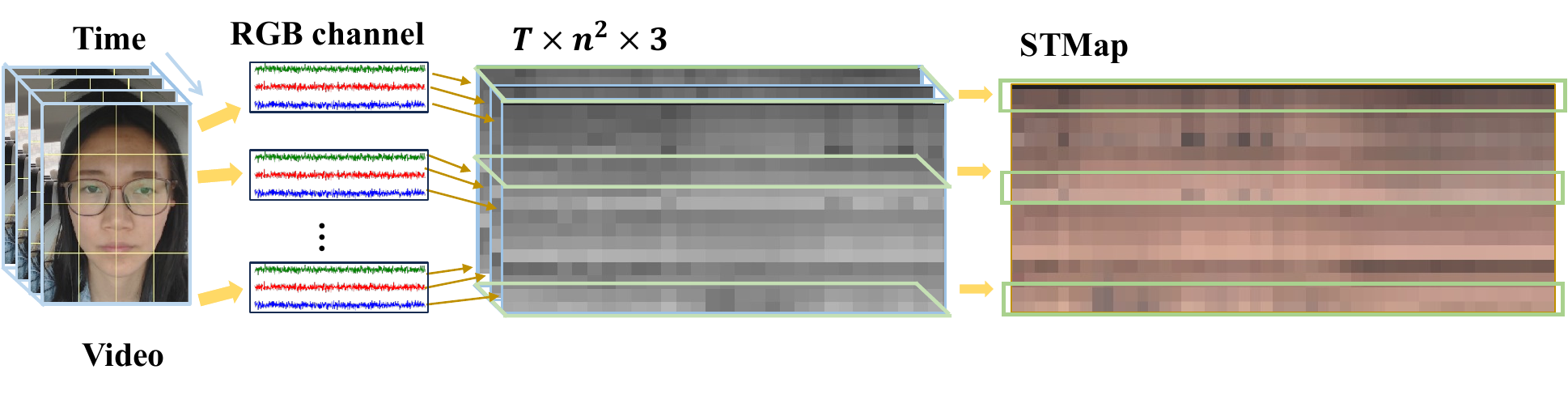}\\
\end{center}
\vspace{-3mm}
\caption{Illustration of producing STMap from RGB video. }\label{stmap}
\vspace{-3mm}
\end{figure*}

As shown in Figure \ref{stmap}, first, we utilized MediaPipe Face Mesh \citep{kartynnik2019real} to perform landmark detection on each frame, obtaining \(468\) facial key points. Next, we define the width of the facial region as the horizontal distance between the boundary points of the outer cheeks, and the height as \(1.2\) times the vertical distance between the chin and the eyebrow center, thus cropping the entire facial region. Then, we apply an HSV color range threshold to segment the skin area, effectively removing non-facial regions such as the eyes and background. Specifically, we convert the region of interest (ROI) to the HSV color space, and define the skin color range by setting the hue (H) between \(0\) and \(20\), the saturation (S) between \(20\) and \(255\), and the value (V) between \(70\) and \(255\). 
Finally, we generate the STMap from the skin regions extracted in each frame of the video. Specifically, for a video with T frames, we divide the skin region in each frame into small blocks of size \((n^2 = 16)\), and compute the average pixel values of the RGB color channels for each block. These average pixel values are then concatenated in the order of the video frames, resulting in an STMap with the shape \((T,16,3)\).

\begin{table}[ht]
\setlength{\tabcolsep}{1pt}
\caption{\textbf{Intra-dataset.} HR estimation performance of RPM methods on RGB, NIR, and mmWave modalities.}\label{app:intra_full}
\centering
\begin{tabular}{l l c c c}
        \toprule[1pt]
        \textbf{Modality} & \textbf{Method} & MAE$\downarrow$ & RMSE$\downarrow$ & P$\uparrow$ \\
        \midrule
        \parbox[t]{2mm}{\multirow{6}{*}{RGB}}
        & CHROM \citep{de2013robust}         & 12.23 & 15.97 & 0.11 \\
        & POS \citep{wang2016algorithmic}     & 12.42 & 16.15 & 0.10 \\
        & GREEN \citep{verkruysse2008remote}     & 14.09 & 17.87 & 0.02 \\
        & ICA  \citep{poh2010advancements}        & 13.31 & 16.93 & 0.06 \\
        & SiNC \citep{speth2023non}          & 13.49 & 16.57 & 0.03 \\
        & Contrast-Phys+ \citep{sun2024contrast} & 15.93 & 18.40 & 0.01 \\
        \midrule
        \parbox[t]{2mm}{\multirow{7}{*}{Video}}
        & DeepPhys \citep{chen2018deepphys}  & 11.97 & 13.17 & 0.20 \\
        & PhysNet \citep{yu2019remote}        &  6.29 &  8.93 & 0.61 \\
        & PhysFormer\citep{yu2022physformer} &  7.85 & 10.17 & 0.41 \\
        & EfficientPhys\citep{liu2023efficientphys}  & 11.27 & 13.51 & 0.20 \\
        & RhythmFormer \citep{zou2024rhythmformer} &  7.21 &  9.84 & 0.45 \\
        & BVPnet$^*$ \citep{das2021bvpnet}       &  7.95 & 10.70 & 0.28 \\
        & RhythmNet$^*$ \citep{niu2019rhythmnet}  &  6.84 &  9.01 & 0.58 \\
        \midrule
        \parbox[t]{2mm}{\multirow{3}{*}{NIR}}
        & DeepPhys \citep{chen2018deepphys} & 15.61 & 17.89 & 0.03 \\
        & PhysNet \citep{yu2019remote}        & 10.69 & 13.21 & 0.12 \\
        & Contrast-Phys+ \citep{sun2024contrast} & 13.65 & 16.08 & 0.05 \\
        \midrule
        \parbox[t]{2mm}{\multirow{3}{*}{mmWave}}
        & IQ-MVED \citep{khan2022contactless} & 14.17 & 18.21 & 0.45 \\
        & VitaNet \citep{khan2022contactless} &  4.94 &  7.15 & 0.94 \\
        & mmFormer \citep{hu2024contactless}  &  3.65 &  5.09 & 0.97 \\
        \bottomrule[1.5pt]
\end{tabular}
\end{table}

\section{Other Results of Evaluation} \label{app:b}

\subsection{Intra-dataset Evaluation}
We trained baseline methods using RGB, NIR, and millimeter-wave modalities on our dataset and validated the results. As shown in Table \ref{app:intra_full}, our results closely match those from similar datasets, particularly for the DL method, yielding reasonable performance. This effectively validates the physiological signals in our dataset (e.g., ECG and BVP) as input modalities and the correlation between input and output.

\subsection{Cross-dataset Evaluation: Trained on Other Datasets}
We conducted cross-dataset training, with results shown in Table \ref{app:cross_full}. We observed that when trained on simpler datasets such as PURE and UBFC-rPPG, traditional methods (e.g., CHROM, POS) performed comparably to the DL methods on our dataset. Notably, the DL method using STMap outperformed the direct video input approach.

\begin{table}[ht]
\centering
\caption{\textbf{Cross-dataset.} HR estimation performance of baselines with different training sets.}\label{app:cross_full}
\begin{tabular}{l l c c c}
  \toprule[1pt]
  \textbf{Method} & \textbf{Train Set} & MAE$\downarrow$ & RMSE$\downarrow$ & P$\uparrow$ \\
  \midrule
  CHROM \citep{de2013robust}  & N/A         & 12.23 & 15.97 & 0.11 \\
  POS \citep{wang2016algorithmic}  & N/A         & 12.42 & 16.15 & 0.10 \\
  ICA \citep{poh2010advancements}  & N/A         & 13.31 & 16.93 & 0.06 \\
  \midrule
  \multirow{3}{*}{SiNC \citep{speth2023non}}      & PURE        & 11.46 & 15.01 & 0.11 \\
                             & UBFC-rPPG   & 14.26 & 18.00 & 0.12 \\
                             & PhysDrive   & 6.29  & 8.93  & 0.61 \\
  \midrule
  \multirow{3}{*}{PhysNet \citep{yu2019remote}}   & PURE        & 13.52 & 16.81 & 0.05 \\
                             & UBFC-rPPG   & 14.68 & 17.98 & 0.02 \\
                             & PhysDrive   & 7.85  & 10.17 & 0.41 \\
  \midrule
  \multirow{3}{*}{RhythmFormer \citep{zou2024rhythmformer}} & PURE     & 11.72 & 15.13 & 0.11 \\
                             & UBFC-rPPG   & 12.53 & 15.87 & 0.08 \\
                             & PhysDrive   & 7.95  & 10.70 & 0.28 \\
  \midrule
  \multirow{3}{*}{BVPnet$^*$ \citep{das2021bvpnet}}     & PURE      & 10.59 & 14.11 & 0.11 \\
                             & UBFC-rPPG   & 10.46 & 13.72 & 0.12 \\
                             & PhysDrive   & 6.84  & 9.01  & 0.58 \\
  \midrule
  \multirow{3}{*}{RhythmNet$^*$ \citep{niu2019rhythmnet}}  & PURE      &  9.30 & 12.21 & 0.14 \\
                             & UBFC-rPPG   &  9.86 & 13.20 & 0.14 \\
                             & PhysDrive   &  7.21 &  9.84 & 0.45 \\
  \midrule
  \multirow{3}{*}{DeepPhys \citep{chen2018deepphys}} & PURE     & 18.57 & 21.67 & 0.01 \\
                             & UBFC-rPPG   & 18.74 & 21.87 & 0.02 \\
                             & PhysDrive   & 11.97 & 13.17 & 0.20 \\
  \midrule
  \multirow{3}{*}{Physformer \citep{yu2022physformer}}  & PURE   & 12.93 & 16.16 & 0.06 \\
                             & UBFC-rPPG   & 14.34 & 17.49 & 0.03 \\
                             & PhysDrive   & 7.85  & 10.17 & 0.41 \\
  \midrule
  \multirow{3}{*}{EfficientPhys \citep{liu2023efficientphys}} & PURE & 17.59 & 20.71 & 0.03 \\
                             & UBFC-rPPG   & 17.42 & 20.56 & 0.00 \\
                             & PhysDrive   & 11.27 & 13.51 & 0.20 \\
  \bottomrule[1.5pt]
\end{tabular}

\end{table}

\subsection{Cross-dataset Evaluation: Tested on Other Datasets}\label{Cross-dataset Evaluation: Tested on Other Datasets}
We also trained the DL method on PhysDrive and tested it on the PURE and UBFC-rPPG datasets. The results are displayed in Table \ref{app:cross_test}. We observed that the performance of the same baseline on PURE and UBFC-rPPG was better. This indicates that our dataset can be effectively used for training DL models and further applied in other scenarios. It also highlights that the training difficulty of PhysDrive is greater than that of PURE and UBFC-rPPG, which is one of the primary reasons for introducing this dataset.

\begin{table*}[ht]
\addtocounter{table}{-1}  
  \setlength{\tabcolsep}{3pt}
  \small
  \centering
    \refstepcounter{table}\label{app:cross_test}
    \centering
    \caption{\textbf{Cross-dataset.} HR estimation performance of baselines when trained in PhysDrive and tested on PURE and UBFC-rPPG.}
  \adjustbox{max width=\textwidth}{
\begin{tabular}{l *{6}{c}}
  \toprule[1.5pt]
   \textbf{Test Set} & \multicolumn{3}{c}{\textbf{PURE}}
    & \multicolumn{3}{c}{\textbf{UBFC-rPPG}} \\
  \cmidrule(lr){2-4}\cmidrule(lr){5-7}
   \textbf{Method} & MAE$\downarrow$ & RMSE$\downarrow$ & P$\uparrow$
    & MAE$\downarrow$ & RMSE$\downarrow$ & P$\uparrow$ \\
  \midrule
  CHROM  \citep{de2013robust}        &  9.79 & 12.76 & 0.37 &  7.23 &  8.92 & 0.51 \\
  POS \citep{wang2016algorithmic}           &  9.82 & 13.44 & 0.34 &  7.35 &  8.04 & 0.49 \\
  SiNC \citep{speth2023non}          & 18.33 & 21.89 & 0.14 & 18.34 & 21.89 & 0.13 \\
  PhysNet \citep{yu2019remote}       & 15.99 & 19.40 & 0.08 & 12.84 & 15.84 & 0.13 \\
  RhythmFormer \citep{zou2024rhythmformer}   & 14.19 & 18.07 & 0.12 & 13.64 & 16.81 & 0.11 \\
  BVPnet$^*$ \citep{das2021bvpnet}         & 14.10 & 17.45 & 0.13 & 12.59 & 15.37 & 0.13 \\
  RhythmNet$^*$ \citep{niu2019rhythmnet}      & 13.76 & 17.41 & 0.16 & 10.01 & 14.35 & 0.20 \\
  \bottomrule[1.5pt]
      \end{tabular}}\\[1pt]

\end{table*}

\begin{table*}[!t]
\setlength{\tabcolsep}{1.5pt}
\caption{\small
\textbf{Cross-dataset Scenario Evaluation with Traditional Methods.} HR estimation performance of RGB‐based baselines under varying lighting and motion conditions.}
\label{cross_scenario_traditional}
\scriptsize
\begin{center}
\adjustbox{max width=\textwidth}{
\begin{tabular}{l 
  *{7}{c c}}
  \toprule[1.5pt]
  \multirow{2}{*}{\textbf{Method}}
    & \multicolumn{2}{c}{\textbf{E.M.\&D.}}
    & \multicolumn{2}{c}{\textbf{Noon}}
    & \multicolumn{2}{c}{\textbf{Night}}
    & \multicolumn{2}{c}{\textbf{R.\&C.}}
    & \multicolumn{2}{c}{\textbf{Stationary}}
    & \multicolumn{2}{c}{\textbf{Talking}}
    & \multicolumn{2}{c}{\textbf{All}} \\
  \cmidrule(lr){2-3}\cmidrule(lr){4-5}\cmidrule(lr){6-7}
  \cmidrule(lr){8-9}\cmidrule(lr){10-11}\cmidrule(lr){12-13}\cmidrule(lr){14-15}
    & MAE$\downarrow$ & P$\uparrow$
    & MAE$\downarrow$ & P$\uparrow$
    & MAE$\downarrow$ & P$\uparrow$
    & MAE$\downarrow$ & P$\uparrow$
    & MAE$\downarrow$ & P$\uparrow$
    & MAE$\downarrow$ & P$\uparrow$
    & MAE$\downarrow$ & P$\uparrow$ \\
  \midrule
  CHROM  \citep{de2013robust}
    & 11.79 & 0.10 & 10.27 & 0.26 & 14.37 & -0.01
    & 12.25 & 0.07 & 12.45 & 0.13 & 12.35 & 0.07 & 12.36 & 0.12 \\
  POS    \citep{wang2016algorithmic}
    & 12.05 & 0.07 & 10.44 & 0.24 & 14.36 & -0.02
    & 12.22 & 0.09 & 12.69 & 0.10 & 12.20 & 0.06 & 12.79 & 0.07 \\
  GREEN  \citep{verkruysse2008remote}
    & 13.56 & 0.01 & 13.53 & 0.03 & 15.16 & -0.02
    & 13.73 & 0.02 & 14.13 & 0.05 & 13.94 & 0.01 & 14.28 & 0.02 \\
  ICA    \citep{poh2010advancements}
    & 13.38 & 0.01 & 12.57 & 0.06 & 14.62 & -0.03
    & 13.00 & 0.03 & 13.86 & 0.03 & 13.41 & 0.04 & 13.73 & 0.03 \\
  \bottomrule[1.5pt]
\end{tabular}}
\end{center}
\vspace{-0.3cm}
\end{table*}

\subsection{Cross-dataset Evaluation on Different Scenarios} \label{Cross-dataset Evaluation on Different Scenarios}
We compared the performance variations of traditional and DL methods under different lighting and motion scenarios, as shown in Table \ref{cross_scenario} and Table \ref{cross_scenario_traditional}. All methods reached their best performance around noon and had the lowest performance at night. Additionally, although the brightness on cloudy and rainy days fluctuated, it remained stable, and the performance was better than during the morning and evening when lighting fluctuated with the driving direction. As expected, the performance of all methods decreased when the driver engaged in additional conversation.
\begin{table*}[!t]
\setlength{\tabcolsep}{1.5pt}
\caption{\small
\textbf{Cross-dataset Scenario Evaluation When Trained on Other Datasets.} HR estimation performance of RGB vision–based baselines under varying lighting and motion conditions when trained on PURE and UBFC.}
\label{cross_scenario}
\scriptsize
\begin{center}
\adjustbox{max width=\textwidth}{
\begin{tabular}{l l *{7}{c c}}
  \toprule[1.5pt]
  \multirow{2}{*}{\textbf{Method}} & \textbf{Condition}
    & \multicolumn{2}{c}{\textbf{E.M.\&D.}} 
    & \multicolumn{2}{c}{\textbf{Noon}} 
    & \multicolumn{2}{c}{\textbf{Night}} 
    & \multicolumn{2}{c}{\textbf{R.\&C.}} 
    & \multicolumn{2}{c}{\textbf{Stationary}} 
    & \multicolumn{2}{c}{\textbf{Talking}} 
    & \multicolumn{2}{c}{\textbf{All}} \\
  \cmidrule(lr){3-4}\cmidrule(lr){5-6}\cmidrule(lr){7-8}
  \cmidrule(lr){9-10}\cmidrule(lr){11-12}\cmidrule(lr){13-14}\cmidrule(lr){15-16}
  & 
   \textbf{Train Set} & MAE$\downarrow$ & P$\uparrow$ 
    & MAE$\downarrow$ & P$\uparrow$ 
    & MAE$\downarrow$ & P$\uparrow$ 
    & MAE$\downarrow$ & P$\uparrow$ 
    & MAE$\downarrow$ & P$\uparrow$ 
    & MAE$\downarrow$ & P$\uparrow$ 
    & MAE$\downarrow$ & P$\uparrow$ \\
  \midrule
  \multirow{2}{*}{SiNC \citep{speth2023non}}
    & PURE     & 11.16 & 0.06 & 10.06 & 0.17 & 13.24 & 0.01 & 11.39 & 0.11 & 11.90 & 0.11 & 11.02 & 0.13 & 11.46 & 0.11 \\
    & UBFC     & 15.99 & 0.03 & 11.62 & 0.26 & 15.42 & -0.01 & 13.99 & 0.08 & 14.07 & 0.14 & 14.39 & 0.08 & 14.26 & 0.12 \\
  \midrule
  \multirow{2}{*}{DeepPhys \citep{chen2018deepphys}}
    & PURE     & 22.73 & -0.03 & 17.18 & 0.00 & 16.96 & -0.02 & 18.19 & 0.03 & 18.71 & 0.00 & 18.58 & 0.00 & 18.57 & 0.01 \\
    & UBFC     & 22.05 & 0.02  & 17.16 & 0.02 & 16.83 & -0.01 & 18.32 & 0.04 & 18.47 & 0.03 & 18.60 & 0.03 & 18.74 & 0.02 \\
  \midrule
  \multirow{2}{*}{PhysNet \citep{yu2019remote}}
    & PURE     & 15.06 & 0.05 & 11.99 & 0.09 & 13.76 & -0.01 & 13.30 & 0.03 & 13.76 & 0.05 & 13.34 & 0.03 & 13.52 & 0.05 \\
    & UBFC     & 16.92 & 0.03 & 12.79 & 0.07 & 13.59 & 0.04  & 14.42 & 0.02 & 14.50 & 0.06 & 14.33 & -0.01& 14.68 & 0.02 \\
  \midrule
  \multirow{2}{*}{Physformer \citep{yu2022physformer}}
    & PURE     & 14.55 & 0.03 & 10.98 & 0.16 & 12.82 & 0.02 & 12.18 & 0.07 & 12.90 & 0.12 & 12.53 & 0.07 & 12.93 & 0.06 \\
    & UBFC     & 16.39 & 0.05 & 12.76 & 0.06 & 13.22 & 0.03 & 13.57 & 0.03 & 14.51 & 0.05 & 14.09 & 0.01 & 14.34 & 0.03 \\
  \midrule
  \multirow{2}{*}{EfficientPhys \citep{liu2023efficientphys}}
    & PURE     & 21.46 & -0.03 & 15.39 & 0.03 & 15.86 & 0.00 & 17.06 & 0.00 & 17.73 & 0.03 & 17.44 & -0.02& 17.59 & 0.03 \\
    & UBFC     & 21.25 & 0.02  & 15.61 & 0.03 & 15.79 & -0.01& 17.26 & -0.02& 17.82 & 0.02 & 17.44 & 0.00 & 17.42 & 0.00 \\
  \midrule
  \multirow{2}{*}{RhythmFormer \citep{zou2024rhythmformer}}
    & PURE     & 12.51 & 0.04 & 10.40 & 0.19 & 13.65 & -0.05& 11.73 & 0.04 & 11.45 & 0.13 & 12.09 & 0.07 & 11.72 & 0.11 \\
    & UBFC     & 12.98 & 0.01 & 11.04 & 0.15 & 12.94 & 0.05 & 12.18 & -0.01& 12.16 & 0.11 & 12.57 & 0.05 & 12.53 & 0.08 \\
  \midrule
  \multirow{2}{*}{BVPnet$^*$ \citep{das2021bvpnet}}
    & PURE     &  6.72 & 0.08 & 10.47 & 0.14 & 13.94 & -0.04& 10.32 & 0.05 & 10.89 & -0.02& 10.29 & -0.02& 10.59 & 0.11 \\
    & UBFC     & 13.80 & 0.01 &  9.99 & 0.16 & 11.14 & -0.06& 10.74 & 0.08 & 11.85 & -0.01& 11.08 & -0.03& 10.46 & 0.12 \\
  \midrule
  \multirow{2}{*}{RhythmNet$^*$ \citep{niu2019rhythmnet}}
    & PURE     &  7.91 & 0.04 &  8.87 & 0.23 & 11.96 & -0.06&  8.40 & 0.14 &  9.61 & 0.15&  8.99 & 0.11&  9.30 & 0.14 \\
    & UBFC     &  6.83 & 0.03 & 10.56 & 0.15 & 12.25 & -0.03&  9.95 & 0.01 & 10.18 & 0.11&  9.56 & 0.17&  9.86 & 0.14 \\
  \bottomrule[1.5pt]
\end{tabular}}
\end{center}
\vspace{-0.3cm}
\end{table*}

\end{document}